\documentclass[10pt,twocolumn,letterpaper]{article}

\usepackage{cvpr}
\usepackage{times}
\usepackage{epsfig}
\usepackage{graphicx}
\usepackage{amsmath}
\usepackage{amssymb}
\usepackage{multirow}
\usepackage{verbatim}
\newcommand*\rot{\rotatebox{90}}
\newcommand\norm[1]{\left\lVert#1\right\rVert}
\usepackage{algorithm}
\usepackage[noend]{algpseudocode}
\cvprfinalcopy 

\def\cvprPaperID{****} 

\begin{document}

\title{Dynamic Representations Toward Efficient Inference on Deep Neural Networks by Decision Gates}

\author{
  Mohammad Saeed Shafiee\thanks{The work has been done while as a visiting scholar at University of Waterloo.} \\
  Faculty of Electrical Engineering\\
  Shahid Rajee University\\
  Tehran, Iran\\
  \texttt{ms.shafiee@srttu.edu} \\
  \and
  Mohammad Javad Shafiee, Alexander Wong \\
  Vision \& Image Processing Research Group\\
  Waterloo AI Institute, University of Waterloo \\
  DarwinAI Corp., Waterloo, Ontario, Canada\\
  \texttt{\{mjshafiee,a28wong\}@uwaterloo.ca} \\
}
\maketitle

\begin{abstract}
    While deep neural networks extract rich features from the input data, the current trade-off between depth and computational cost makes it difficult to adopt deep neural networks for many industrial applications, especially when computing power is limited. Here, we are inspired by the idea that, while deeper embeddings are needed to discriminate difficult samples (i.e., fine-grained discrimination), a large number of samples can be well discriminated via much shallower embeddings (i.e., coarse-grained discrimination). In this study, we introduce the simple yet effective concept of \textbf{decision gates} (d-gate), modules trained to decide whether a sample needs to be projected into a deeper embedding or if an early prediction can be made at the d-gate, thus enabling the computation of dynamic representations at different depths.  The proposed d-gate modules can be integrated with any deep neural network and reduces the average computational cost of the deep neural networks while maintaining modeling accuracy. The proposed d-gate framework is examined via different network architectures and datasets, with
    experimental results showing that leveraging the proposed d-gate modules led to a $\sim$43\% speed-up and 44\% FLOPs reduction on ResNet-101 and 55\% speed-up and 39\% FLOPs reduction on DenseNet-201 trained on the \mbox{CIFAR10} dataset with only $\sim$2\% drop in accuracy. Furthermore, experiments where d-gate modules are integrated into ResNet-101 trained on the ImageNet dataset demonstrate that it is possible to reduce the computational cost of the network by 1.5 GFLOPs without any drop in the modeling accuracy.
\end{abstract}

\vspace{-0.25cm}
\section{Introduction}
\vspace{-0.25cm}
Advances in deep learning have led to tremendous success in a wide variety of applications in visual and audio perception such as image classification~\cite{szegedy2016rethinking,he2016deep,huang2017densely}, object detection~\cite{he2017mask,redmon2017yolo9000,liu2016ssd}, and speech recognition~\cite{amodei2016deep}. What's interesting about deep convolutional neural networks is that it brings together the notions of feature extraction, feature projection, and prediction within an end-to-end learning framework to produce more coherent and more discriminative predictors.

The excitement around deep learning and recent findings that increasing network depth~\cite{simonyan2014very} typically results in greater modeling capacity has led researchers to focus on designing deeper and more complex deep neural networks to improve modeling accuracy.

Although having deeper architectures was demonstrated to provide better modeling performances, a number of challenges arise as we increase network depth.  Besides becoming more prone to overfitting and becoming more difficult to train to convergence, deeper neural networks also result in a significant increase in not only the number of parameters in the network, but also dramatically increases the computational cost of network inference.

A number of strategies have been proposed to tackle the various challenges associated with deeper neural network architectures.  For example, Szegedy {\it et al.}~\cite{szegedy2015going,szegedy2016rethinking}  introduced the concept of inception modules which helps to increase the depth of deep neural networks while maintaining the number of parameters. Specifically, inception modules consist of several convolutional layers with different receptive field sizes fed by the same inputs. This architecture helps the model to extract better features with fewer number of computations. While this new module mitigates the computational complexity to some extent and extends the possibility of having deeper networks with fewer parameters, deeper network architectures still suffer from vanishing gradient issues and thus a degradation in learning.

He {\it et al.}~\cite{he2016deep} took a different strategy to address the problem and tackled the former issue of degradation in learning deeper neural networks (e.g., vanishing gradient) by introducing the concept of residual learning, where learning is based on the residual mapping rather than directly on the unreferenced mapping. This novel idea brought forth the possibility of much larger and deeper networks by easing the training of such networks.

Following that, Xie~{\it et al.}~\cite{xie2017aggregated} incorporated the idea behind inception modules (i.e., split-transform-merge strategy) within a residual block structure to provide better subspace modeling while resolving the degradation problem at the same time, resulting in a ResNext architecture that achieved improved modeling accuracy. Several other architectures have been proposed based on these observations to provide better modeling accuracy. For example, the DenseNet~\cite{huang2017densely} architecture connects each layer to every other layer in a feed-forward fashion. In this network architecture, the feature maps of all preceding layers are used as inputs to the next layer, thus alleviating the vanishing gradient issue. Zoph~{\it et al.}~\cite{zoph2017learning} devised an evolutionary algorithm to search through a huge set of possible computational blocks and found the most optimized block architecture to design a every deep neural network with. The parameters of the computational block are trained during the search procedure and the whole network are fine-tuned with the training data as well.

The computational cost associated with deep neural networks remain a significant bottleneck for deployment in many industrial applications.  Although some applications can leverage high-performance computing units such as GPUs to enable real-time operation, this comes at a very high financial cost in the form of cloud computing costs or on-premise equipment and power costs.  Furthermore, there are a large number of industrial applications where access to high-performance computing devices is simply not possible.  As such, mechanisms for reducing computational cost of deep neural networks while retaining modeling accuracy is highly desired.

To tackle the issue of computational cost, a wide variety of methods have been proposed.  One common strategy is precision reduction~\cite{jacob2017quantization}, where the data representation of a network is reduced from the typical 32-bit floating point precision to low-bit fixed-point or integer precision. This technique is suitable mainly for the specialized hardware with the faster lower precision arithmetic calculation. Another common strategy is model compression~\cite{han2015deep}, which involves leveraging traditional data compression methods such as weight thresholding, hashing, and Huffman coding. Such compression methods are mostly beneficial in storage reduction unless the hardware used supports accelerated sparse multiplications.  Other strategies include the use of teacher-student strategies~\cite{hinton2015distilling}, where a larger teacher network is used to train a smaller student network, as well as the use of evolutionary algorithms~\cite{javad2016evonet,shafiee2016evolutionary} for evolving the architecture of deep neural networks over generations to be more compact.

More recently, conditional computation~\cite{bengio2015conditional,denoyer2014deep,liu2017dynamic,wu2018blockdrop} and early prediction~\cite{teerapittayanon2016branchynet} methods have been proposed to tackle this issue, which involve the dynamic execution of different modules within a network. Conditional computation methods have largely been motivated by the idea that residual networks can be considered as an ensemble of shallower networks.  As such, these methods take advantage of skip connections to determine which residual modules are necessary to be executed, with most leveraging reinforcement learning.  For example, a controller is typically trained within a reinforcement framework where the controller plays the role of deciding which network block needs to be executed based the input image. The controller is usually a shallower network.

Early prediction techniques, on the other hand, divide the network into several partitions, with fully-connected layers (i.e., classification layers) integrated into the network at the end of each partition.  In previous work~\cite{teerapittayanon2016branchynet}, the new network is then trained based a multi-loss function with respect to all of the integrated fully-connected layers. After the network and all of the integrated fully-connected layers are trained, a threshold is calculated for each of the integrated fully-connected layer classifiers based on the outputs of the Softmax layer to determine whether the sample can be predicted in the current fully-connected layer or requires prediction after the next partition.  A particular limitation to past early prediction approaches is that the fully-connected layers are trained based on a cross-entropy loss, which has been shown to produce predictions that may not be as reliable as desired\footnote{We will discuss more about this issue in Section~\ref{sec:Method}.}.   Furthermore, not only are such previous techniques quite difficult to set up, they also require the network to be trained from scratch with these techniques integrated for strong performance.

In this study, we explore the idea of early prediction but instead draw inspiration from the soft-margin support vector~\cite{cortes1995support} theory for decision-making.  Specifically, we introduce the concept of decision gates (d-gate), modules that are trained via hinge loss to decide whether a sample needs to be projected into a deeper embedding or if an early prediction can be made at the d-gate, thus enabling the conditional computation of dynamic representations at different depths.  The proposed d-gate modules can be integrated with any deep neural network without the need to train networks from scratch, and thus reduces the average computational complexity of the deep neural networks while maintaining modeling accuracy.

\section{Methodology}
\label{sec:Method}
Deeper neural network architectures have been demonstrated to provide a better subspace embedding of data when compared to shallower architectures, resulting in a better discrimination of data in the new space and better modeling accuracy. Although it has been demonstrated that going to deeper layers provides better representational features (i.e., more discriminative feature) for the data, particularly for samples with high similarity to each other that would require fine-grained discrimination, the hypothesis here is that a huge set of samples in the data space do not need a very detailed and descriptive feature representation to be discriminated from other samples (i.e., coarse-grained discrimination).  As such, these samples can be predicted earlier at shallower embeddings, which would lead to faster processing time.

Interestingly, this hypothesis has been explored in relation to the human brain, where the processing time of an individual when discriminating between objects that are highly distinctive based on coarse-grained visual characteristics is much faster than when discriminating between more difficult objects where the differentiating characteristics are more subtle and fine-grained~\cite{mace2009time,kheradpisheh2016humans}. For example, Mac{\'e} {\it et al.}~\cite{mace2009time} studied the  human processing speed when categorizing natural scenes as containing either an animal (superordinate level), or a specific animal (bird or dog, basic level). What they found was that the human visual system can very rapidly recognize superordinate categories of objects by accessing a coarse/abstract visual representation, while the processing time was much slower for discriminating between objects at the basic level since  additional time is needed to visually analyze  more detailed representations of the objects.

\begin{figure}
 \vspace{-0.55cm}
\setlength{\tabcolsep}{1pt}
\begin{tabular}{cccccccc}
    &\multicolumn{3}{c}{Cat} & \multicolumn{3}{c}{Truck}\\
     \rot{\hspace{0.22 cm}\tiny Classified} &\includegraphics[]{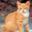}&\includegraphics[]{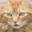}&\includegraphics[]{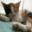}& \includegraphics[]{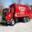}&\includegraphics[]{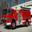}&\includegraphics[]{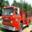}\\

     \rot{\hspace{0.1 cm}\tiny Misclassified}&\includegraphics[]{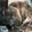}&\includegraphics[]{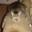}  &\includegraphics[]{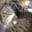}&\includegraphics[]{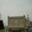}&\includegraphics[]{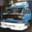}&\includegraphics[]{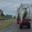}\\
     {\tiny  Predicted Label} & Frog & Dog & Bird & Ship & Car & Plane
\end{tabular}
    \caption{Example classification results using shallow embeddings, with the first row illustrating correctly classified samples and the bottom row illustrating incorrectly classified samples.  It can be observed that the samples that were correctly classified have highly discriminative characteristics that humans can recognize very rapidly.  On the other hand, the samples that were incorrectly classified have very subtle characteristics that even humans have a hard time picking up and require longer processing for humans to recognize.}
    \label{fig:exp1}
    \vspace{-0.55cm}
\end{figure}

To better motivate the hypothesis, Figure~\ref{fig:exp1} illustrates a simple experiment  where we investigate the type of classification results that can be produced using shallow embeddings.  In particular, the output features from the first three residual blocks of ResNet-101~\cite{he2016deep} (which constitutes a shallow embedding) are utilized to train a linear classifier for the \mbox{CIFAR10} dataset, such that the correctly classified samples as well as misclassified samples are identified and analyzed qualitatively.  It can be observed that the samples that were correctly classified have highly discriminative characteristics that humans can recognize very rapidly.  On the other hand, the samples that were incorrectly classified have very subtle characteristics that even humans have a hard time picking up and require longer processing for humans to recognize. The interesting observation here is that the misclassified images are assigned a class label which can be considered in the same superordinate category as their ground truth class labels, which reflects the observations made in the aforementioned study with regards to superordinate-level vs. basic-level discrimination~\cite{mace2009time}. For example, the misclassified cat images are classified as frog, dog, or bird (which including cats are all types of animals) while the misclassified truck images are categorized as ship, car, or plane (which including trucks are all types of vehicles).

Inspired by the result in Figure~\ref{fig:exp1}, we make a hypothesis that while deeper subspace embeddings are necessary to discriminate samples that lie close to the decision boundaries in the lower embedding space, their effect on samples that already lie far from decision boundaries in the shallower embedding space may be insignificant and unnecessary.
Therefore, an effective yet efficient mechanism for determining the distance between samples and the decision boundaries in the lower layers of the network would make it possible to perform early predictions on these samples without projecting them into a deeper embedding space. Such an approach would reduce the average computational cost of prediction significantly.

\subsection{Soft-Margin Support Vector Theory}
\label{sec:softmargin}

Despite the promise shown in early prediction mechanisms~\cite{teerapittayanon2016branchynet}, devising an efficient yet effective way to determine whether a sample is a boundary sample is a very challenging problem, especially in the conventional setup applied to train a deep neural network. Although cross-entropy loss with Softmax~\cite{goodfellow2016deep} is the most common approach to train deep neural networks, past research~\cite{liang2017soft} have argued that cross-entropy loss provides a small margin between the decision boundaries and the training data. This issue associated with cross-entropy loss is exemplified by the frequent observation that neural networks could misclassify samples that are just slightly different from the training data, and even minor modifications to a sample can lead to a change in prediction.  As such, this limitation also makes it very difficult to determine whether a sample is an boundary sample or not based on the output of Softmax layer.

To address this limitation, we are instead motivated to draw inspiration from soft-margin support vector~\cite{cortes1995support} theory for decision-making, where we aim to optimize with the notion of maximizing the soft-margin of the separating hyperplane in scenarios where the data is not perfectly separable.  More specifically, a training sample with a higher margin would receive less weight than an example with lower margin.  As a result, for the purposes of early prediction at shallower embeddings, a soft-margin loss function can provide the best possible linear hyperplane to determine samples that are likely to be classified correctly at an earlier stage.

\begin{figure}
    \centering
    \includegraphics[width=5cm]{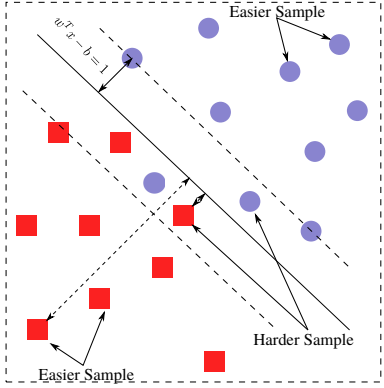}
    \caption{Example of hinge loss within a binary classification scenario. Loosely speaking, optimization via a hinge loss positions the samples in a way that samples that are harder to classify are closer to the decision boundary while easier samples are further from the decision boundary. }
    \label{fig:hinge_loss_boundary}
    \vspace{-0.25cm}
\end{figure}

One effective way to maximize the classification margin based on soft-margin support vector theory is the use of a hinge loss. In the binary classification case, the hinge loss is used to determine the decision boundary that maximizes the margin between the samples of two classes:
\begin{align}
    \ell(y) = \max \big(0, 1- y \cdot \hat{y} \big) \hspace{0.5 cm} \text{s.t.} \hspace{0.5 cm} \hat{y} = w^Tx-b
    \label{eq:hinge}
\end{align}
where $y$ is the ground truth label for the input data $x$ and $\hat{y}$ is the predicted class label via the  d-gate module with the set of weights $w$ and biases $b$.  The set of weights $w$ has a dimensionality of $f \times c$, where $f$ denotes the number of input features and $c$ denotes the number of class labels in the classification task. This formulation provides an important benefit where the result of $w^Tx-b$ is the distances of the sample to the corresponding decision boundary of each class label in the embedding space. As shown in Figure~\ref{fig:hinge_loss_boundary} the interesting property of this loss function would be the harder samples to classify is closer to the decision boundary compared to the easier samples which their distances can be computed easily  by \eqref{eq:hinge}. Hinge loss can be extended to multi-class problems in an one-vs-all manner; it is not differentiable but the sub-gradient is applied with respect to $w$ to optimize the loss function:
\begin{align}
\frac{\partial l}{\partial w} = \begin{cases}
-y \cdot x &\text{$\hat{y} < 1$}\\
0 &\text{otherwise}
\end{cases}
\vspace{-0.25cm}
\end{align}
when $y$ and $\hat{y}$ have the same sign and $|\hat{y}| \geq 1$, the loss is zero and, therefore, there is no gradient; otherwise the loss increases linearly. Therefore, traditional gradient descent can be adapted here, where a step is taken in the direction of a vector selected from the function's sub-gradient~\cite{shalev2011pegasos} to find the optimized values.

\subsection{Decision Gates}
\label{sec:d_gate}
\begin{figure*}
\vspace{-0.75cm}
    \centering
    \includegraphics[width = 18cm]{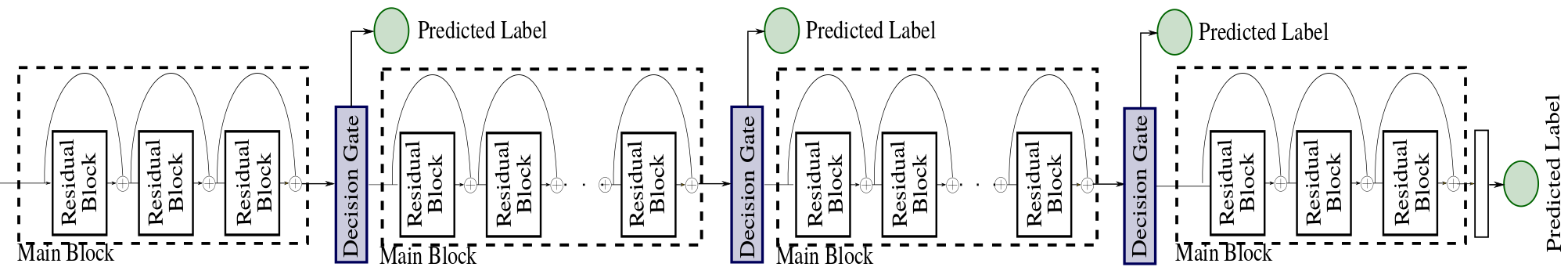}
    \caption{\footnotesize Decision gates (d-gate) are integrated directly into deep neural networks, and are trained to predict whether decisions can be made at the d-gate or require projecting into deep embeddings.}
    \label{fig:reseNet_d-gate}
    \vspace{-0.35cm}
\end{figure*}
Inspired by the aforementioned soft-margin support vector theory, we formulate the early prediction problem as a risk minimization problem, and introduce a set of single-layer feed-forward networks (which we will refer to as \mbox{\textbf{decision gates}} (d-gate)) that are integrated directly into a deep neural network (see Figure~\ref{fig:reseNet_d-gate}).  The goal of d-gate modules is to not only decide whether a sample requires projection into a deep embedding space, but also minimize the risk of early wrong classifications as well.  Specifically, we train d-gate modules that are integrated into a deep neural network via a hinge loss~\cite{dogan2016unified} that minimizes the risk of early misclassification in lower embedding while deciding whether the sample is a boundary sample.

 Training the d-gate module in this way provides a linear classifier where samples that do not require deeper embeddings for discrimination are those with larger distances (i.e., with the positive sign) from the decision boundary.  It is important to note that single layer nature of d-gate modules is designed to account for efficiency.

The d-gate module is trained via the training data utilized to train the deep neural network and the objective for each d-gate module is to minimize the classification error on the training data. Therefore, the loss function on the training data can be formulated as:
\begin{align}
  \mathcal{L}(Y,\hat{Y};w,b) = \frac{1}{n} \sum_{i=1}^n \max\Big(0,1 - y_i(w^Tx_i-b)\Big) - \lambda \norm{w} \nonumber
\end{align}
where $Y$ and $\hat{Y}$ denote the set of ground truth labels and predicted ones for all training data. What is most interesting about $\mathcal{L}(Y,\hat{Y};w,b)$ is the fact that $\mathcal{L}(\cdot)$ is a convex function of $w$ and $b$, and as such can be optimized via gradient descent.  As a result, the d-gate can be trained within a mini-batch training framework, which makes it is very convenient for utilizing in the training of deep neural networks with large datasets.

\begin{algorithm}[t]
\caption{Decision Gate}\label{alg:d-gate}
\begin{algorithmic}[1]
\Procedure{d-gate}{$x,t$}\Comment{{\tiny $x$: feature set, $t$: d-gate threshold }}
\State $\bar{d}\gets w^T x - b$  \Comment{{\tiny $w$: weights, $b$: bias }} \\
                         \Comment{{\tiny $\bar{d}$: the set of distances to the boundaries}}

\If{$\max(\bar{d}) \geq t$}\Comment{{\tiny distance to a boundary is larger than the threshold}}
\State $cl \gets \arg\max({\bar{d}})$
\State \textbf{return} $cl$\Comment{{\tiny$cl$: class label is predicted.} }
\Else
\State \textbf{return} $NULL$\Comment{{\tiny The sample is projected to a deeper embedding space.}}
\EndIf
\EndProcedure
\end{algorithmic}
\end{algorithm}
\vspace{-0.25cm}
\subsection{Early Prediction Framework}
\label{sec:e_prediction}
Given that the proposed d-gate module can calculate the distance of each sample to the decision boundaries based on $w^Tx-b$, where $x$ is the feature set fed into the d-gate module, it is possible to integrate a d-gate module at any place in the network to enable early prediction.  As such, deciding where to integrate d-gate modules becomes an important consideration, as integrating multiple d-gate modules will allow for early prediction at various embedding depths.

Fortunately, many of the popular network architectures proposed in literature and in widespread use are compositional architectures consisting of several well-formed computational blocks. Therefore, one can take advantage of the modularity of such network architectures to integrate d-gate modules after each computational block in the network. For example, as seen in Figure~\ref{fig:reseNet_d-gate}, the ResNet-101 architecture is composed of four computational blocks (main blocks) of 3, 4, 23, and 3 residual blocks, respectively, and the d-gate modules are applied after each of the first three computational blocks.

The calculated distances are compared to the decision threshold $t$ of each d-gate to determine whether early prediction on the sample can be performed at the d-gate or have the sample moved to a deeper stage of the deep neural network to project into a better embedding space for improved prediction. Algorithm~\ref{alg:d-gate} demonstrates how a d-gate module determines whether the sample should be projected to a deeper embedding space (i.e., returns NULL) or the class label can be predicted correctly with a high probability at the current d-gate module.

The samples that are far from the decision boundaries result to output larger values in $w^Tx-b$; therefore, if the d-gate distance for a sample satisfies the d-gate decision threshold, the class corresponding to the largest distance is assigned as the predicted class label for the sample in this early prediction step.
\vspace{-0.2 cm}

\section{Experimental Results \& Discussion}
\label{sec:Exp}
We explore and investigate the efficacy of the proposed d-gate modules using a series of experiments across different network architectures for multiple datasets.

More specifically, the first experiment studies the effect of leveraging hinge loss vs. cross entropy loss in the training of d-gate modules. Following that, the second experiment studies the trade-offs between efficiency vs. effectiveness when varying the decision thresholds of d-gate modules integrated into a ResNet-101 network architecture.  The third experiment studies the performance of integrating d-gate modules (in terms of computational cost and modeling accuracy) into two well-known network architectures: i) ResNet-101, and ii) DenseNet-201.  Finally, the fourth experiment studies in a statistical as well as qualitative fashion the distribution of classes that is handled for early prediction by the individual d-gate modules integrated within a ResNet-101 architecture.

\subsection{Datasets}
 To train different network architectures to analyze the advantage of the proposed d-gate modules, two datasets were utilized: i) CIFAR-10~\cite{krizhevsky2009learning}, and ii) ImageNet~\cite{imagenet_cvpr09}. The CIFAR-10 dataset is comprised of $32\times32$ natural images with 50000 training images and 10000 test images in 10 different class labels. The ImageNet dataset has 1000 class labels of natural images. For this paper we use the 1.2M training data of ILSVRC2012 challenge dataset for training the network while the validation set comprised of 50000 images are used to test the networks.
\vspace{-0.4 cm}
\subsubsection*{Training Setup}
\vspace{-0.25 cm}
Each d-gate module in the network is trained independently based on all training data. This is because of the fact that if the d-gate modules are trained together, then the d-gate modules toward the end of the network might not be provided with enough training data as the samples are classified in the earlier d-gate modules in the network. It is worth noting that the parameters of the convolutional layers are frozen during the training of the d-gate modules, and only a linear classifier is trained for calculating the distance of sample to the decision boundary.

The reported decision threshold values used by the d-gate modules in the paper were calculated via a cross-validation approach, where the goal is to set the decision thresholds such that the resulted network achieves a pre-determined accuracy (with specific margins from the original accuracy). Although the d-gate modules were trained separately, the decision thresholds were selected in a sequential manner to incorporate the dependency among the d-gate modules in the network. More specifically, the decision thresholds are tuned via an objective function such that a specific modeling accuracy is achieved at the lowest average FLOPs computation based on the validation dataset.

\subsection{Experiment 1: Cross Entropy Vs. Hinge Loss}

Given that one of the main contributions and key differentiating aspects of the proposed work is the introduction of a hinge loss that minimizes the risk of early misclassification in lower embeddings to train the d-gate modules within a deep neural network, we first explore the efficacy of such an approach compared to the utilization of cross-entropy loss for d-gate training.  Although cross-entropy loss with Softmax is the most commonly-used approach to training deep neural networks, past studies~\cite{NIPS2018_7364,liang2017soft} have argued that it results in a small margin between the decision boundaries and the training data.  Softmax trained via Cross Entropy does not provide as much information about the distance of the sample to the margin when compared to trained via the Hinge loss approach. To demonstrate the effectiveness of the hinge loss leveraged in the proposed d-gate modules compared to the cross-entropy loss, a comparative experiment was conducted with adding two decision gates to ResNet-101 trained on CIFAR-10 dataset. However, rather than train using the proposed hinge loss, the decision gates were instead trained via a cross-entropy loss.  This enables us to compare the effect of hinge loss vs. cross-entropy loss on decision gate functionality.

\begin{figure}
\vspace{-0.25cm}
  \centering
    \includegraphics[width=7cm]{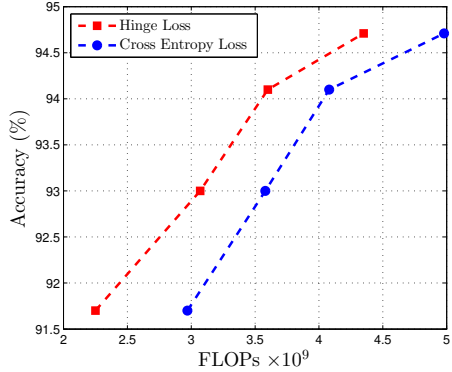}
    \caption{ Accuracy vs. number of FLOPs: The performance of the network with decision gates trained via the proposed hinge loss is compared with decision gates trained via a conventional cross-entropy approach. It can be observed that the decision gates trained via the hinge loss provide greater computational efficiency with higher accuracy than if cross-entropy loss is leveraged.
    }
    \label{fig:cross-ent-hinge}
    \vspace{-0.65cm}
\end{figure}

Figure~\ref{fig:cross-ent-hinge} demonstrates the accuracy vs. number of FLOPs for the network where the decision gates were trained based on the proposed hinge loss approach compared to trained using a regular cross-entropy training procedure.  It can be observed that, with the same number of FLOPs in the network, the network where the decision gates were trained based on the proposed hinge loss provides much higher modeling accuracy compared to that trained via cross-entropy loss. The accuracy gap increases exponentially when the decision gates are configured such that the network uses fewer number of FLOPs.  What this illustrates is the aforementioned issue with the use of cross-entropy loss and decision boundaries.  Therefore, it can be clearly observed that leveraging the proposed hinge loss for decisions gates yields significant performance benefits over the use of a conventional cross-entropy loss.

\subsection{Experiment 2: Efficiency Vs. Effectiveness}
\begin{figure}
\vspace{-0.25cm}
    \centering
    \begin{tabular}{cc}
    \setlength{\tabcolsep}{1pt}
         \includegraphics[width = 4cm]{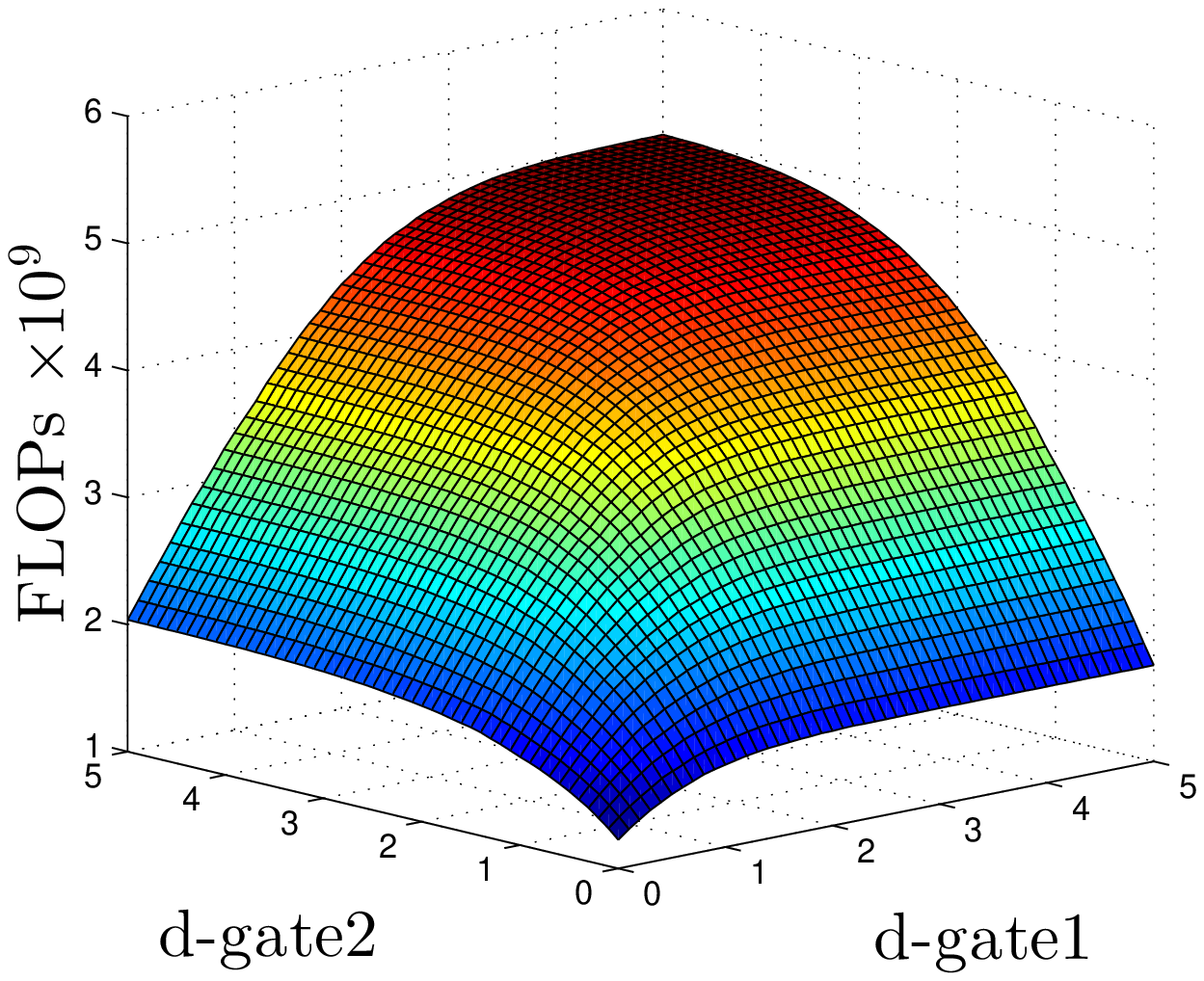}& \includegraphics[width = 4cm]{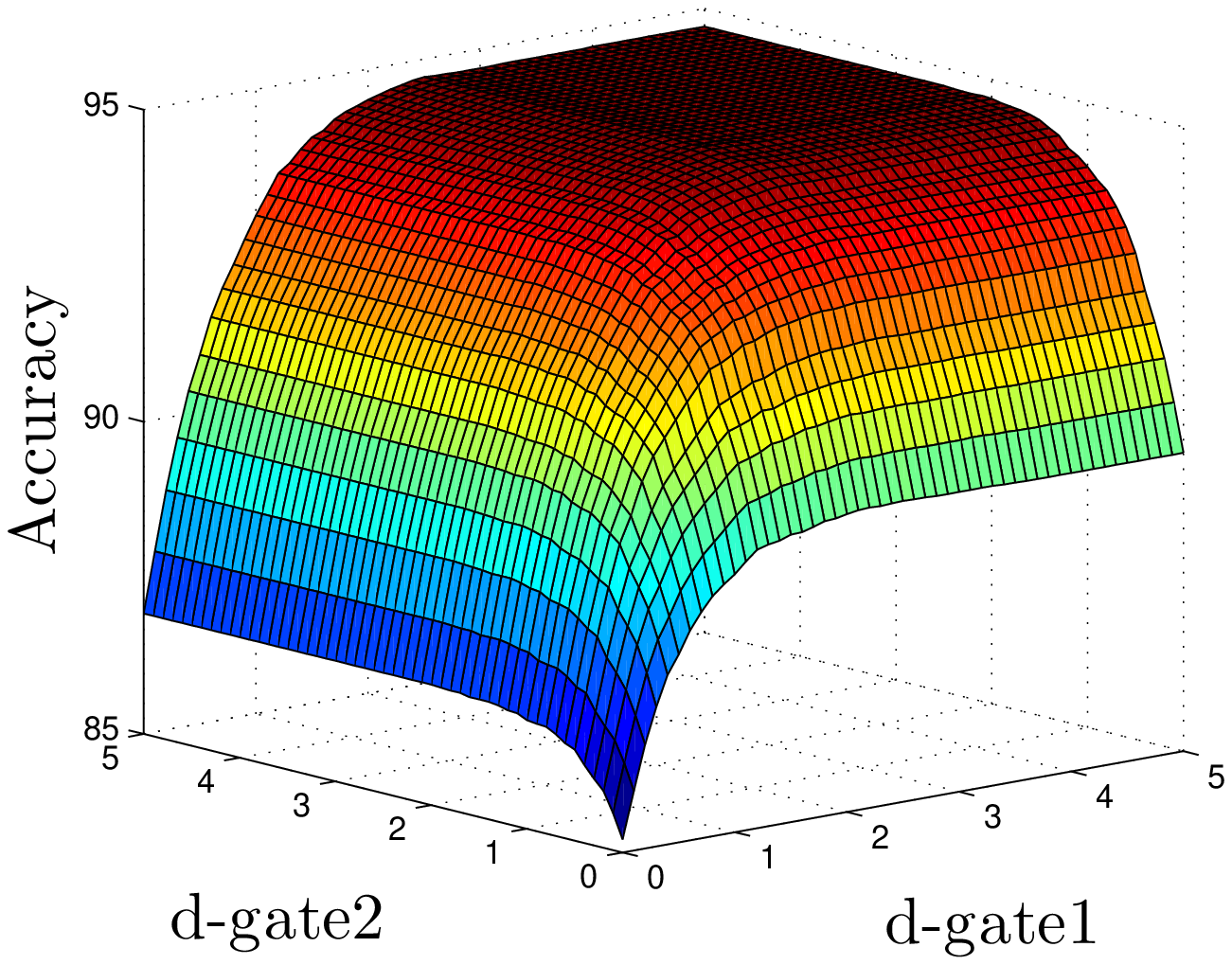}\\
         (a)&(b)
    \end{tabular}
    \caption{Effectiveness Vs. efficiency trade-off using d-gate modules: The plots show how changing the decision thresholds of the d-gate modules can affect the efficiency of the network (a) and the overall modeling accuracy of the network (b).  Therefore, the incorporation of d-gate modules into network architectures result in networks with dynamic representations that can achieve dynamic trade-offs between speed and accuracy as needed.  }
    \label{fig:d-gate_flop_accuracy}
    \vspace{-0.55cm}
\end{figure}

In the second experiment, we analyze the effect of incorporating d-gate modules on computational complexity and accuracy of a deep neural network.  More specifically, two d-gate modules were added to a ResNet-101 architecture trained on the CIFAR-10 dataset after the first and second main blocks, as shown in Figure~\ref{fig:reseNet_d-gate}. The first main block in ResNet-101 is comprised of three residual blocks, while the second main block has four residual blocks. We then examined the trade-offs between efficiency and effectiveness of the resulting network by varying the  decision thresholds of the d-gate modules.
Figure~\ref{fig:d-gate_flop_accuracy}-(a) illustrates the number of FLOPs when the d-gate decision thresholds are varied. It can be observed that, as expected, when the d-gate thresholds are decreased the number of FLOPs decreases as well, since more samples undergo early prediction at shallower embeddings. Figure~\ref{fig:d-gate_flop_accuracy}-(b) illustrates the modeling accuracy when the d-gate decision thresholds are varied. It can be seen that, when the d-gate decision thresholds are set to $(t1,t2)=(0,0)$, the number of FLOPs is reduced by 6$\times$ compared to the original network with a 9\% accuracy reduction. However the modeling accuracy increases by specifying higher threshold for the d-gate modules and it reaches to the same level of the accuracy as the original network when the decision thresholds exceed 2, while still providing a $\sim$20\% reduction in the number of FLOPs compared to the original network. The plots in Figure~\ref{fig:d-gate_flop_accuracy} illustrates that, by incorporating the proposed d-gate modules within a network architecture, it is possible to construct networks with dynamic representations that can achieve dynamic trade-offs between speed and accuracy.  This dynamic nature of networks with integrated d-gate modules can be especially important for applications where when the amount of computing power varies from time to time due to factors such as energy consumption.
\vspace{-0.25cm}
\subsection{Experiment 3: D-gate Performance}
The efficacy of the proposed d-gate modules is examined with two different network architectures (ResNet-101~\cite{he2016deep} and DenseNet201~\cite{huang2017densely}) on the CIFAR10 dataset in this section. A key benefit of the proposed d-gate modules is that it enables fine control over the trade-off between modeling accuracy and computational cost by adjusting the d-gate decision thresholds. As mentioned in the previous section, by decreasing the d-gate decision thresholds, the number of samples undergoing early prediction increases, thus reducing the average computational cost of network predictions greatly.  For this experiment, we integrated three d-gate modules in ResNet-101 (after the first and second main blocks and third main blocks) and DenseNet-201 (after the first, second and third dense blocks), and explore different d-gate configurations.  The networks are implemented in the Pytorch framework and the prediction speeds are reported based on single Nvidia Titan Xp GPU.

\begin{table*}[h]
\vspace{-0.4 in}
    \centering
    \caption{ Experimental results for ResNet-101 with different d-gate configurations. The average number of FLOPs and accuracy for each configuration are compared with that of the original networks. $dg(t1,t2,t3)$ denotes network with three d-gate modules configured with decision thresholds $t1$, $t2$, and $t3$ respectively. }
\begin{tabular}{l|c|c|c|c|c}
        \bf Model &  \bf ACC. & \bf FLOPs (G) & \bf FLOPs Reduction &\bf Speed (ms) &  \bf Speed-up     \\\hline
        dg(1.0, 0.6,1.5)& 91.99\%  & 2.22   &   55.77\%  & 6.39 & 55.71\% \\
        dg(1.0, 1.8,1.5)& 92.97\%  & 2.82   &   43.82\%  & 8.18 &43.31\% \\
        dg(1.5, 1.7,1.0)& 93.99\%  & 3.20   &   36.25\%  & 9.24 & 35.96\%\\
        dg(2.4, 2.4,1.5)& 94.70\%  & 3.93   &   21.71\%  & 11.23& 22.17\%  \\ \hline
       Original      & 94.71\%  & 5.02   & --            & 14.43& --
    \end{tabular}
    \label{tab:tab1_ResNet-101}
\end{table*}

Table~\ref{tab:tab1_ResNet-101}  shows the experimental results for ResNet-101 with 4 different d-gate configurations compared with the original ResNet-101 accuracy and performance.
It can be observed from Table~\ref{tab:tab1_ResNet-101} that the computational cost of ResNet network can be reduced by 1090 MFLOPs while maintaining the same level of accuracy as to the original ResNet-101 by integrating three d-gate modules with decision thresholds of $(t1,t2,t3)=(2.4,2.4,1.5)$ for the d-gate modules which results to 22\% speed up on the model performance. The integration of d-gate modules can reduce the computational cost of ResNet-101 network by 1.78$\times$ (i.e., lower by 2.2 GFLOPs) with $\sim$1.7\% drop in accuracy compared to the original ResNet-101 (with distance thresholds \mbox{$(t1,t2,t3)=(1.0,1.8,1.5)$} in d-gate1, d-gate2, d-gate3), resulting in a $\sim$43\% speed-up on the GPU.

The experimental results for DenseNet-201 are shown in Table~\ref{tab:tab1_DenseNet}. As seen, it is possible to reduce the number of FLOPs by 460 MFLOPs without any meaningful drop in accuracy, leading to a 28\% speed-up on the GPU. Furthermore, a 2$\times$ speed-up can be achieved with d-gate modules, \mbox{$(t1,t2,t3) = (1.1,1.7,1.0)$}, compared to the original DenseNet-201 with a $\sim$2\% accuracy margin.

The experimental results illustrates that the proposed \mbox{d-gate} modules can lead to a significant increase in prediction speed with the ability to control the accuracy drop in a model which makes it well-suited for industrial applications.

\begin{table*}[h]
    \centering
    \caption{ Experimental results for DenseNet-201 with different d-gate configurations. The average number of FLOPs and accuracy for each configuration are compared with that of the original networks. $dg(t1,t2,t3)$ denotes network with three d-gate modules configured with decision thresholds $t1$, $t2$, and $t3$ respectively.}
\begin{tabular}{l|c|c|c|c|c}
        \bf Model &  \bf ACC. & \bf FLOPs (G) & \bf FLOPs Reduction &\bf Speed (ms) &  \bf Speed-up     \\\hline
        dg(1.0, 1.0,1.0)& 92.21\%  & 1.50   &   44.85\%  & 7.27 & 61.45\% \\
        dg(1.1, 1.7,1.0)& 93.20\%  & 1.66   &   38.97\%  & 8.39 & 55.51\%\\
        dg(1.5, 2.4,1.0)& 94.20\%  & 1.85   &   31.98\%  & 9.82 & 47.93\%\\
        dg(3.5, 3.5,2.5)& 95.20\%  & 2.26   &   16.91\%  & 13.56 &  28.10\%   \\ \hline
       Original      & 95.29\%  & 2.72   &    --      &  18.86 & --
    \end{tabular}
    \label{tab:tab1_DenseNet}
\end{table*}

\begin{figure}[t]
\vspace{-0.1in}
    \centering
    \setlength{\tabcolsep}{1pt}
    \begin{tabular}{cccc} \includegraphics[width = 1.1cm]{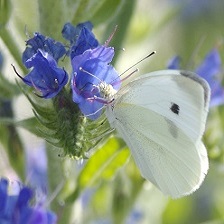} &\includegraphics[width = 1.1cm]{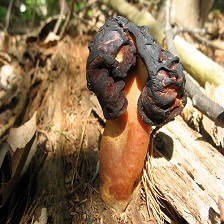} &\includegraphics[width = 1.1cm]{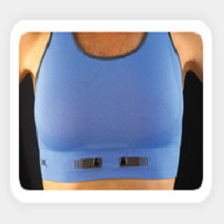}&\includegraphics[width = 1.1cm]{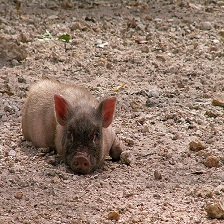}  \\ \includegraphics[width = 1.1cm]{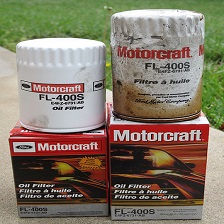} &\includegraphics[width = 1.1cm]{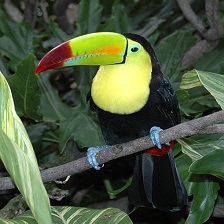} &\includegraphics[width = 1.1cm]{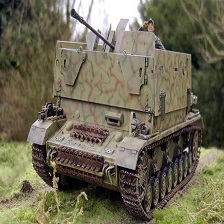}&\includegraphics[width = 1.1cm]{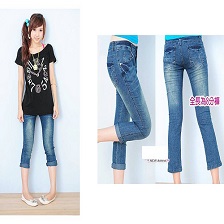}  \\
    \end{tabular}
    \caption{Image examples from classes where early prediction is performed at d-gate1 and d-gate2. It can be observed that d-gate1 and d-gate2 are very good at discriminating classes with very distinguishable characteristics than other classes, where shallower embedding are sufficient to make good classification predictions.}
  \label{fig:d-gate_Correct_classfied}
  \vspace{-0.35cm}
\end{figure}

\begin{figure*}[!h]
\vspace{-0.45cm}
    \centering
     \setlength{\tabcolsep}{1pt}
    \begin{tabular}{ccc}
         \includegraphics[width=6cm]{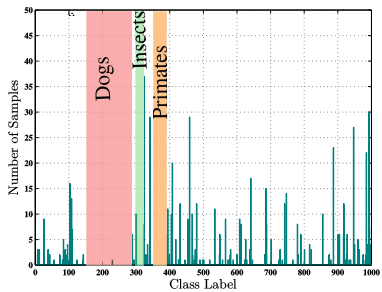}&
         \includegraphics[width=6cm]{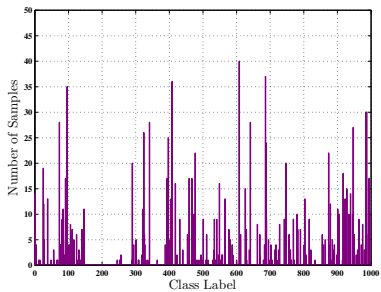}&
         \includegraphics[width=6cm]{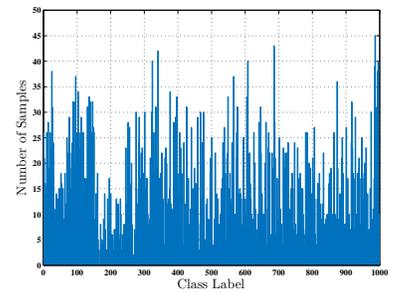}\\
         (a) d-gate1 & (b) d-gate2 & (c) d-gate3
    \end{tabular}
    \caption{Prediction statistics at each of the three d-gate modules integrated into ResNet-101 (configured as $(t1,t2,t3) = (1.0,1.0,1.0)$) for the ImageNet test dataset.}
    \label{fig:d-gate_stats_images}
    \vspace{-0.65cm}
\end{figure*}

Additionally, the performance of the proposed d-gate framework is examined on ResNet-101 trained on the ImageNet dataset~\cite{imagenet_cvpr09}.  For comparison purposes, three state-of-the-art methods were also evaluated: i) \textbf{BlockDrop~\cite{wu2018blockdrop}}, which learns to dynamically choose which layers of a deep neural network to execute during inference via a policy network, ii) \textbf{ACT~\cite{figurnov2017spatially}}, which utilizes an RNN with halting unit to determine the the probability of progressing with the computation, and iii) \textbf{SACT~\cite{figurnov2017spatially}}, which extends upon ACT to apply to each spatial position of multiple image blocks.

The Top-1 accuracy of ResNet-101 trained on ImageNet was reported~\cite{wu2018blockdrop} as 76.4\% at 15.6~GFLOPs\footnote{The comparison results are reported from~\cite{wu2018blockdrop}. }. ACT~\cite{figurnov2017spatially} achieved an accuracy of 75.3\% at 13.4~GFLOPs, while SACT~\cite{figurnov2017spatially} achieved an accuracy of 75.8\% at 14.4~GFLOPs. BlockDrop, which is considered as the state-of-the-art in this area achieved an accuracy of 76.8\% at 14.7~GFLOPs (5.7\% reduction in computations). The proposed d-gate framework achieved 76.8\% at 14.1~GFLOPs (9.6\% reduction in computations).

\begin{figure}[h]
    \centering
    \setlength{\tabcolsep}{1pt}
    \begin{tabular}{cccccc}
    \includegraphics[width = 1.1cm]{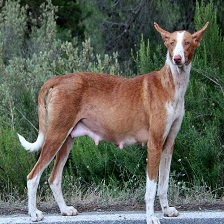} &\includegraphics[width = 1.1cm]{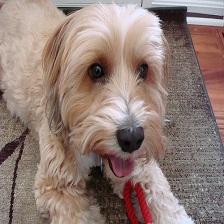} &\includegraphics[width = 1.1cm]{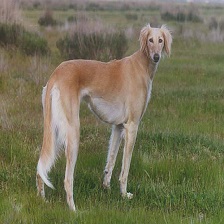}&\includegraphics[width = 1.1cm]{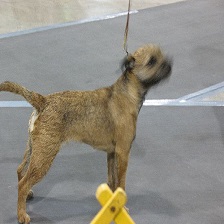} &\includegraphics[width = 1.1cm]{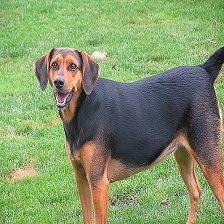}&\includegraphics[width = 1.1cm]{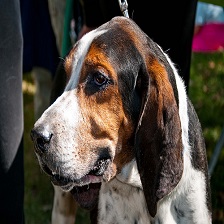} \\
       \includegraphics[width = 1.1cm]{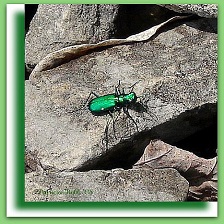} &\includegraphics[width = 1.1cm]{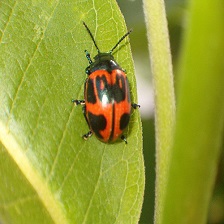} &\includegraphics[width = 1.1cm]{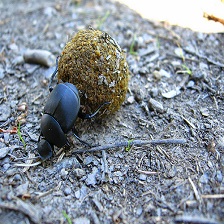}&\includegraphics[width = 1.1cm]{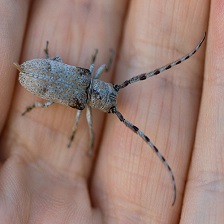} &\includegraphics[width = 1.1cm]{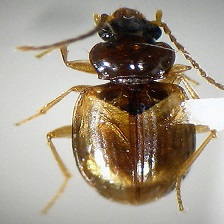}&\includegraphics[width = 1.1cm]{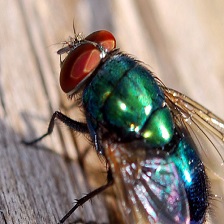} \\
          \includegraphics[width = 1.1cm]{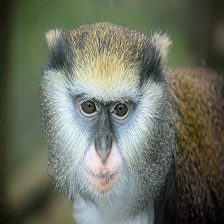} &\includegraphics[width = 1.1cm]{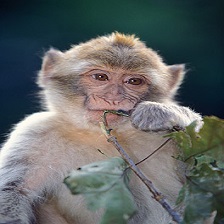} &\includegraphics[width = 1.1cm]{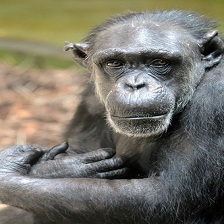}&\includegraphics[width = 1.1cm]{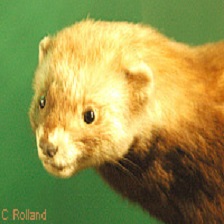} &\includegraphics[width = 1.1cm]{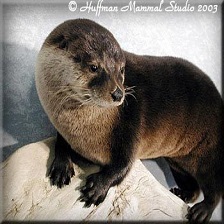}&\includegraphics[width = 1.1cm]{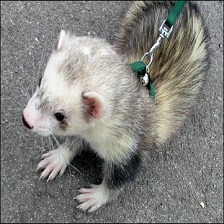}
    \end{tabular}
    \caption{Example images from classes where prediction was performed at d-gate3.  A very interesting observation is that the these classes are typically associated with fine-grained classification problems. For example, classes associated with different breeds of dogs, different types of insects, different types of primates, and different types of weasels, which are distinguishable only by more subtle fine-grained characteristics, were classified at d-gate3, which leverage at deeper embedding, as opposed to at d-gate1 and d-gate2, which leverage shallower embeddings. }
    \label{fig:d-gate_unCorrect_classfied}
    \vspace{-0.56cm}
\end{figure}

\subsection{Experiment 4: Distribution of Classes Across d-gate Modules}
In the fourth and final experiment, we study in a statistical as well as qualitative fashion the distribution of classes that is handled for early prediction by the individual d-gate modules integrated within a ResNet-101 architecture on the classification of the ImageNet validation dataset. More specifically, the main objective of this experiment is to study which of the classes are classified at the different d-gate modules within a network. The plots in Figure~\ref{fig:d-gate_stats_images}(a)-(c) demonstrate the prediction statistics of the validation set of ImageNet dataset at each of the three d-gate modules (d-gate1, d-gate2, d-gate3) that are integrated into ResNet-101, respectively. The X-axis of plots shows the 1000 class labels of ImageNet dataset while Y-axis represents the number of images that can be classified by each d-gate module correctly with a distance equal or larger than one from the decision boundary (i.e., which means the d-gate module classifies them with a very high certainty).

Figure~\ref{fig:d-gate_Correct_classfied} shows image examples from classes where early prediction is performed at d-gate1 and d-gate2. It can be observed that both d-gate1 and d-gate2 are very good at discriminating classes that have very distinguishable characteristics when compared to other classes in the ImageNet dataset.  This observation matches intuition as  highly distinguishable classes can be well characterized and discriminated with shallower embeddings, and as such earlier d-gate modules are sufficient to make good classification predictions given the shallower embeddings.

The other interesting observation in the plot (a) of Figure~\ref{fig:d-gate_stats_images} is that, there are certain class labels which require projection to deeper embeddings for accurate classification, and as such are predicted at a later d-gate module like d-gate3 within the network. We investigated those classes to figure out which type of objects are hard to classify at earlier stages of the network. Figure~\ref{fig:d-gate_unCorrect_classfied} demonstrates three examples of those class labels. As seen, the \mbox{d-gate1} module could not classify different dog breeds, types of primates, types of insects, and types of weasels. Intuitively, a shallow network cannot discriminate against fine-grain class labels in the ImageNet dataset as there are several classes of dog, monkey and insect in the ImageNet dataset. Due this fact, it is suggested that a hierarchical classification approach might help to improve the classification accuracy in the deep neural networks specifically for these types of architectures. It has been also observed by other research~\cite{bagherinezhad2018label} that refining the class labels can improve the classification accuracy of the models.
\vspace{-0.25cm}
\section{Conclusion}
\vspace{-0.25cm}
In this paper we proposed a new framework, (d-gate) modules, to handle the trade-off between the computational complexity and modeling accuracy of deep neural networks. The d-gate modules are added in different parts of the network architecture to decide whether the input data needs to be projected into the deeper embedding space of the network or not. Experimental results showed  the effectiveness of the proposed framework in reducing the computational complexity while maintaining the modeling accuracy. The qualitative experiment  showed that it is possible to apply a hierarchical classification technique to improve the modeling performance as the future work for this framework.
\vspace{-0.25cm}
\section*{Acknowledgement}
\vspace{-0.25cm}
We gratefully acknowledge the support of NVIDIA Corporation with the donation of the Titan Xp GPU used for this research.

{\small 
\bibliographystyle{ieee}
\bibliography{egbib}

\begin{thebibliography}{10}\itemsep=-1pt

\bibitem{amodei2016deep}
D.~Amodei, S.~Ananthanarayanan, R.~Anubhai, J.~Bai, E.~Battenberg, C.~Case,
  J.~Casper, B.~Catanzaro, Q.~Cheng, G.~Chen, et~al.
\newblock Deep speech 2: End-to-end speech recognition in english and mandarin.
\newblock In {\em International Conference on Machine Learning}, pages
  173--182, 2016.

\bibitem{bagherinezhad2018label}
H.~Bagherinezhad, M.~Horton, M.~Rastegari, and A.~Farhadi.
\newblock Label refinery: Improving imagenet classification through label
  progression.
\newblock {\em arXiv preprint arXiv:1805.02641}, 2018.

\bibitem{bengio2015conditional}
E.~Bengio, P.-L. Bacon, J.~Pineau, and D.~Precup.
\newblock Conditional computation in neural networks for faster models.
\newblock {\em arXiv preprint arXiv:1511.06297}, 2015.

\bibitem{cortes1995support}
C.~Cortes and V.~Vapnik.
\newblock Support-vector networks.
\newblock {\em Machine learning}, 20(3):273--297, 1995.

\bibitem{imagenet_cvpr09}
J.~Deng, W.~Dong, R.~Socher, L.-J. Li, K.~Li, and L.~Fei-Fei.
\newblock {ImageNet: A Large-Scale Hierarchical Image Database}.
\newblock In {\em CVPR09}, 2009.

\bibitem{denoyer2014deep}
L.~Denoyer and P.~Gallinari.
\newblock Deep sequential neural network.
\newblock {\em arXiv preprint arXiv:1410.0510}, 2014.

\bibitem{dogan2016unified}
{\"U}.~Dogan, T.~Glasmachers, and C.~Igel.
\newblock A unified view on multi-class support vector classification.
\newblock {\em Journal of Machine Learning Research}, 17(45):1--32, 2016.

\bibitem{NIPS2018_7364}
G.~Elsayed, D.~Krishnan, H.~Mobahi, K.~Regan, and S.~Bengio.
\newblock Large margin deep networks for classification.
\newblock In {\em Advances in Neural Information Processing Systems 31}. 2018.

\bibitem{figurnov2017spatially}
M.~Figurnov, M.~D. Collins, Y.~Zhu, L.~Zhang, J.~Huang, D.~P. Vetrov, and
  R.~Salakhutdinov.
\newblock Spatially adaptive computation time for residual networks.
\newblock In {\em CVPR}, volume~2, page~7, 2017.

\bibitem{goodfellow2016deep}
I.~Goodfellow, Y.~Bengio, A.~Courville, and Y.~Bengio.
\newblock {\em Deep learning}, volume~1.
\newblock MIT press Cambridge, 2016.

\bibitem{han2015deep}
S.~Han, H.~Mao, and W.~J. Dally.
\newblock Deep compression: Compressing deep neural networks with pruning,
  trained quantization and huffman coding.
\newblock {\em arXiv preprint arXiv:1510.00149}, 2015.

\bibitem{he2017mask}
K.~He, G.~Gkioxari, P.~Doll{\'a}r, and R.~Girshick.
\newblock Mask r-cnn.
\newblock In {\em Computer Vision (ICCV), 2017 IEEE International Conference
  on}, pages 2980--2988. IEEE, 2017.

\bibitem{he2016deep}
K.~He, X.~Zhang, S.~Ren, and J.~Sun.
\newblock Deep residual learning for image recognition.
\newblock In {\em Proceedings of the IEEE conference on computer vision and
  pattern recognition}, pages 770--778, 2016.

\bibitem{hinton2015distilling}
G.~Hinton, O.~Vinyals, and J.~Dean.
\newblock Distilling the knowledge in a neural network.
\newblock {\em arXiv preprint arXiv:1503.02531}, 2015.

\bibitem{huang2017densely}
G.~Huang, Z.~Liu, L.~Van Der~Maaten, and K.~Q. Weinberger.
\newblock Densely connected convolutional networks.
\newblock In {\em CVPR}, volume~1, page~3, 2017.

\bibitem{jacob2017quantization}
B.~Jacob, S.~Kligys, B.~Chen, M.~Zhu, M.~Tang, A.~Howard, H.~Adam, and
  D.~Kalenichenko.
\newblock Quantization and training of neural networks for efficient
  integer-arithmetic-only inference.
\newblock {\em arXiv preprint arXiv:1712.05877}, 2017.

\bibitem{kheradpisheh2016humans}
S.~R. Kheradpisheh, M.~Ghodrati, M.~Ganjtabesh, and T.~Masquelier.
\newblock Humans and deep networks largely agree on which kinds of variation
  make object recognition harder.
\newblock {\em Frontiers in computational neuroscience}, 10:92, 2016.

\bibitem{krizhevsky2009learning}
A.~Krizhevsky and G.~Hinton.
\newblock Learning multiple layers of features from tiny images.
\newblock Technical report, Citeseer, 2009.

\bibitem{liang2017soft}
X.~Liang, X.~Wang, Z.~Lei, S.~Liao, and S.~Z. Li.
\newblock Soft-margin softmax for deep classification.
\newblock In {\em International Conference on Neural Information Processing},
  pages 413--421. Springer, 2017.

\bibitem{liu2017dynamic}
L.~Liu and J.~Deng.
\newblock Dynamic deep neural networks: Optimizing accuracy-efficiency
  trade-offs by selective execution.
\newblock {\em arXiv preprint arXiv:1701.00299}, 2017.

\bibitem{liu2016ssd}
W.~Liu, D.~Anguelov, D.~Erhan, C.~Szegedy, S.~Reed, C.-Y. Fu, and A.~C. Berg.
\newblock Ssd: Single shot multibox detector.
\newblock In {\em European conference on computer vision}, pages 21--37.
  Springer, 2016.

\bibitem{mace2009time}
M.~J.-M. Mac{\'e}, O.~R. Joubert, J.-L. Nespoulous, and M.~Fabre-Thorpe.
\newblock The time-course of visual categorizations: you spot the animal faster
  than the bird.
\newblock {\em PloS one}, 4(6):e5927, 2009.

\bibitem{redmon2017yolo9000}
J.~Redmon and A.~Farhadi.
\newblock Yolo9000: better, faster, stronger.
\newblock {\em arXiv preprint}, 2017.

\bibitem{javad2016evonet}
M.~Shafiee, A.~Mishra, and A.~Wong.
\newblock Deep learning with darwin: Evolutionary synthesis of deep neural
  networks.
\newblock {\em arXiv:1606.04393}, 2016.

\bibitem{shafiee2016evolutionary}
M.~Shafiee and A.~Wong.
\newblock Evolutionary synthesis of deep neural networks via synaptic
  cluster-driven genetic encoding.
\newblock In {\em {NIPS} {Workshop}}, 2016.

\bibitem{shalev2011pegasos}
S.~Shalev-Shwartz, Y.~Singer, N.~Srebro, and A.~Cotter.
\newblock Pegasos: Primal estimated sub-gradient solver for svm.
\newblock {\em Mathematical programming}, 127(1):3--30, 2011.

\bibitem{simonyan2014very}
K.~Simonyan and A.~Zisserman.
\newblock Very deep convolutional networks for large-scale image recognition.
\newblock {\em arXiv preprint arXiv:1409.1556}, 2014.

\bibitem{szegedy2015going}
C.~Szegedy, W.~Liu, Y.~Jia, P.~Sermanet, S.~Reed, D.~Anguelov, D.~Erhan,
  V.~Vanhoucke, and A.~Rabinovich.
\newblock Going deeper with convolutions.
\newblock In {\em Proceedings of the IEEE conference on computer vision and
  pattern recognition}, pages 1--9, 2015.

\bibitem{szegedy2016rethinking}
C.~Szegedy, V.~Vanhoucke, S.~Ioffe, J.~Shlens, and Z.~Wojna.
\newblock Rethinking the inception architecture for computer vision.
\newblock In {\em Proceedings of the IEEE conference on computer vision and
  pattern recognition}, pages 2818--2826, 2016.

\bibitem{teerapittayanon2016branchynet}
S.~Teerapittayanon, B.~McDanel, and H.~Kung.
\newblock Branchynet: Fast inference via early exiting from deep neural
  networks.
\newblock In {\em Pattern Recognition (ICPR), 2016 23rd International
  Conference on}, pages 2464--2469. IEEE, 2016.

\bibitem{wu2018blockdrop}
Z.~Wu, T.~Nagarajan, A.~Kumar, S.~Rennie, L.~S. Davis, K.~Grauman, and
  R.~Feris.
\newblock Blockdrop: Dynamic inference paths in residual networks.
\newblock In {\em Proceedings of the IEEE Conference on Computer Vision and
  Pattern Recognition}, pages 8817--8826, 2018.

\bibitem{xie2017aggregated}
S.~Xie, R.~Girshick, P.~Doll{\'a}r, Z.~Tu, and K.~He.
\newblock Aggregated residual transformations for deep neural networks.
\newblock In {\em Computer Vision and Pattern Recognition (CVPR), 2017 IEEE
  Conference on}, pages 5987--5995. IEEE, 2017.

\bibitem{zoph2017learning}
B.~Zoph, V.~Vasudevan, J.~Shlens, and Q.~V. Le.
\newblock Learning transferable architectures for scalable image recognition.
\newblock {\em arXiv preprint arXiv:1707.07012}, 2(6), 2017.

\end{thebibliography}
}

\end{document}